\documentclass{article}

\usepackage[preprint, nonatbib]{nips_2018}

\usepackage[utf8]{inputenc} 
\usepackage[T1]{fontenc}    
\usepackage{hyperref}       
\usepackage{url}            
\usepackage{booktabs}       
\usepackage{amsfonts}       
\usepackage{nicefrac}       
\usepackage{microtype}      

\usepackage{listings}
\usepackage{graphicx}
\usepackage{siunitx}

\usepackage[font=footnotesize,labelfont=footnotesize]{subcaption}
\usepackage[font=footnotesize,labelfont=footnotesize]{caption}

\usepackage{amsmath}
\usepackage{color}

\definecolor{dkgreen}{rgb}{0,0.6,0}
\definecolor{gray}{rgb}{0.5,0.5,0.5}
\definecolor{mauve}{rgb}{0.58,0,0.82}

\graphicspath{ {images/} }

\title{Learning Configuration Space Belief Model from Collision Checks for Motion Planning}

\author{
  Sumit Kumar\thanks{This work was carried out as an internship at Carnegie Mellon University. Contact: sumit.sks4[at]gmail.com}\\
  \And
  Shushman Choudhary \\
  \AND
  Siddhartha Srinivasa\\
}

\begin{document}

\maketitle

\begin{abstract}
For motion planning in high dimensional configuration spaces, a significant computational bottleneck is collision detection. Our aim is to reduce the expected number of collision checks by creating a belief model of the configuration space using results from collision tests. We assume the robot's configuration space to be a continuous ambient space whereby neighbouring points tend to share the same collision state. This enables us to formulate a probabilistic model that assigns to unevaluated configurations a belief estimate of being collision-free. We have presented a detailed comparative analysis of various kNN methods and distance metrics used to evaluate C-space belief. We have also proposed a weighting matrix in C-space to improve the performance of kNN methods. Moreover, we have proposed a topological method that exploits the higher order structure of the C-space to generate a belief model. Our results indicate that our proposed topological method outperforms kNN methods by achieving higher model accuracy while being computationally efficient. 
\end{abstract}

\section{Introduction}
Motion planning can be defined as the task of finding a continuous path between a start configuration $S$ and a goal configuration $G$ while avoiding any collision with obstacles present in the environment. The robot and obstacle geometry are expressed in 2D or 3D workspace while the path is represented in robot's configuration space (C-space). Robotic motion planning requires exploration of C-space. As the dimensionality of C-space increases, complete exploration becomes computationally intractable. 

For higher dimensional C-space, sampling-based planning algorithms are considered state-of-the-art. It constructs a connectivity graph that implicitly represents the structure of C-space, thus minimizing exploration. The main computations involved in path planning are generating C-space samples, joining nearby collision-free samples and evaluating the collision state (in $C_{free}$ or in $C_{obs}$) of a configuration or a local path by a collision detection module. The latter step, performing a collision check, provides exact information about the configuration but comes at the cost of increased computational load making even the sampling based techniques intractable sometimes. One way to solve the problem is to use C-space belief as a heuristic to guide the search for feasible paths. C-space belief which refers to the belief of unevaluated configurations to be collision-free is evaluated with the help of a probabilistic model of the robot's world.

A number of different nearest neighbour methods and distance metrics have been used by researchers so far to generate the belief model. We have performed a detailed comparative analysis of these in the same validation environment to create a benchmark. To the best of our knowledge, however, all the traditional kNN methods work in the higher dimensional C-space and do not utilize any embedded topological information. We have projected the C-space to a lower dimensional space and proposed a topological method in this projected space to generate the belief model. In the end, we have also presented a comparative analysis of  our topological method and the traditional nearest neighbour ones present in the literature.  

\section{Related Work}

Path planning algorithms have widely used graph-based and sampling-based techniques. Graph-based planning algorithms like Dijkstra's algorithm~\cite{dijkstra1959note}, A*~\cite{hart1968formal} and D*-Lite~\cite{koenig2002d} are used mostly in low dimensional C-spaces where explicit representation is computationally feasible. On the other hand, sampling based techniques are generally applied in high dimensional C-spaces where explicit representation becomes infeasible. Probabilistic Roadmap~\cite{kavraki1994probabilistic} (PRM) planner constructs a graph or a roadmap of sampled configurations and tries to find a path from start to goal state. Rapidly-exploring Random Tree~\cite{lavalle1998rapidly} (RRT) planner expands a tree from start state and terminates after reaching the goal state.   

The original probabilistic roadmap approach explicitly evaluated all nodes and edges of the roadmap for feasibility of path planning. As a result of the expensive nature of collision-checking, the idea of reducing the number of collision checks by generating a probabilistic model of the configuration space and estimating the collision probability of configurations and paths became popular. Recent works have been done on creating a belief model of the C-space and using the probability of collision as a heuristic to guide the search over paths to obtain a feasible path~\cite{nielsen2000two}. Burns et al. incrementally constructs and refines an approximate statistical model of the entire configuration space~\cite{burns2005sampling}. The model is used to estimate the feasibility of an edge in the roadmap reducing the need to invoke collision checker. Pan et al.~\cite{pan2013faster} improves the performance of sample-based motion planners by learning from prior instances. They perform instance-based learning on the approximate C-space and compute collision probability. Choudhary et al.~\cite{choudhury2016pareto} builds up an incremental model of the configuration space using data from previous collision tests and uses this model to search for successively shorter paths that are likely to be feasible. 

For robots with large number of DOFs, planning in configuration space can often be computationally expensive. To speed up the computation, one approach is to split the planning problem into two lower-dimensional sub-problems - planning for the shoulder and elbow joints and that for the wrist joints ~\cite{gochev2014motion}. Researchers have often projected the configuration space to a lower dimensional space and used topological methods for path planning. Pokorny et al.~\cite{pokorny2016high} shows that under some constraints the topological properties are preserved upon projection.

\section{Problem Definition}
We formulate an inexpensive probabilistic model $\Phi$ that takes in a configuration and outputs its belief of being collision-free. The belief depends on the model and its parameters $\Theta$. It is a mapping from robot's configuration space $C$ to a set $S = (-1,1)$. A positive value predicts the query configuration to be in $C_{obs}$ while a negative value estimates it to be in $C_{free}$, higher the magnitude higher is the corresponding probability. Mathematically, this mapping can be expressed as $B(\Phi,\Theta):C \mapsto S$. 

The performance of the model $\Phi$ is evaluated using two metrics: accuracy and average error. Accuracy refers to the fraction of number of query configurations for whom the model correctly predicts its collision state. Average error refers to the mean L2 deviation between the actual collision state and the predicted collision probability of all the query configurations. 


  

\section{Approach}

The robotic platform used in our work is HERB (Home Exploring Robot Butler) at Personal Robotics Lab, Carnegie Mellon University. HERB is a bimanual mobile manipulator comprising of two 7 DOF Barret WAM arms on a Segway base equipped with a suite of image and range sensors. The C-space mentioned in this project is the 7-dimensional joint space of HERB's right arm. We assume that the C-space is a continuous ambient space whereby neighbouring points tend to share the same collision state, i.e., either both or neither of them are in $C_{free}$.

\begin{figure}[ht]
    \begin{minipage}{.32\textwidth}
    \includegraphics[width=\textwidth]{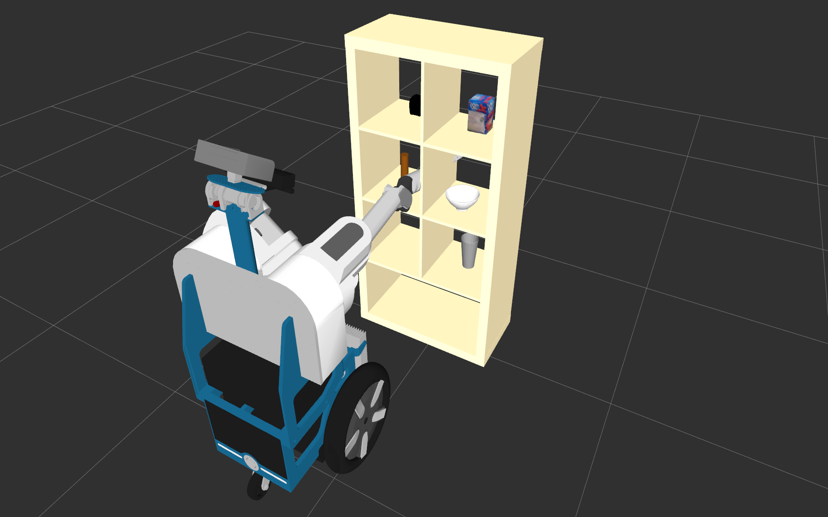}
    \subcaption{shelf}
    \end{minipage}\hfill
    \begin{minipage}{.32\textwidth}
    \includegraphics[width=\textwidth]{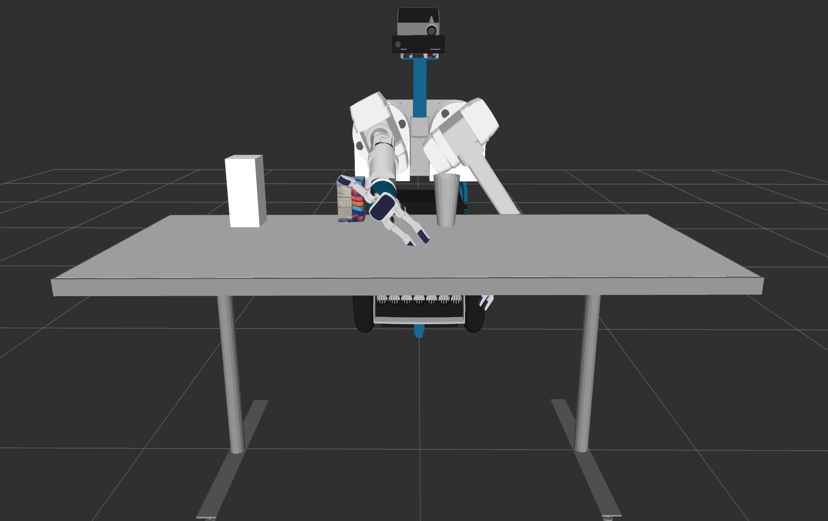}
    \subcaption{table}
    \end{minipage}\hfill
    \begin{minipage}{.32\textwidth}
    \includegraphics[width=\textwidth]{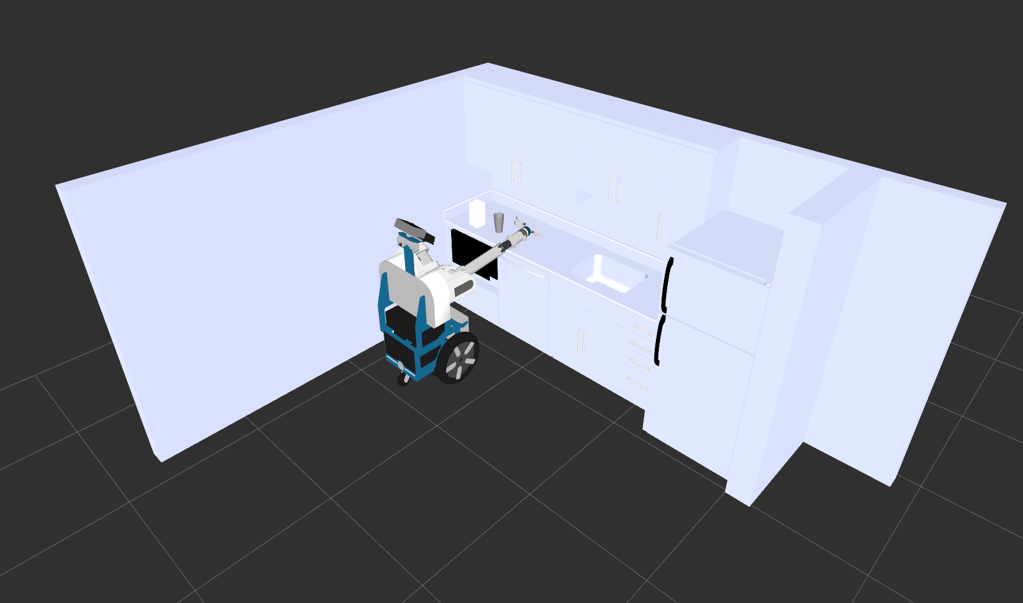}
    \subcaption{kitchen}
    \end{minipage}
\caption{ Typical environments HERB works in.}
\label{fig:envs}
\end{figure}

\subsection{Sobol Sequences}
We created 8 environments typical to what HERB will find itself in everyday life like a kitchen, a room with shelf and table. 3 of these are shown in Figure \ref{fig:envs}. In each of these environments, we generated a training set of $N$ (model parameter) samples using Sobol sequences. The motivation behind preferring Sobol sequences over randomly generated ones is that the former are quasi-random low discrepancy sequences, hence, cover the space more uniformly and leave less number of voids. A Sobol sequence \textbf{S} is a 7D vector whose each element is a non-negative number less than or equal to 1. It is then linearly transformed to the robot's C-space as follows: 
$$\mathbf{q} = \mathbf{L} + (\mathbf{U}-\mathbf{L}) \circ \mathbf{S} $$

where, \textbf{q} is the configuration in C-space, \textbf{L} and \textbf{U} are the vectors of minimum and maximum values of the arm joint angles respectively. Operator $\circ$ carries out element wise multiplication of the two matrices. 

\begin{figure}[ht]
    \begin{minipage}{.5\textwidth}
    \includegraphics[width=\textwidth]{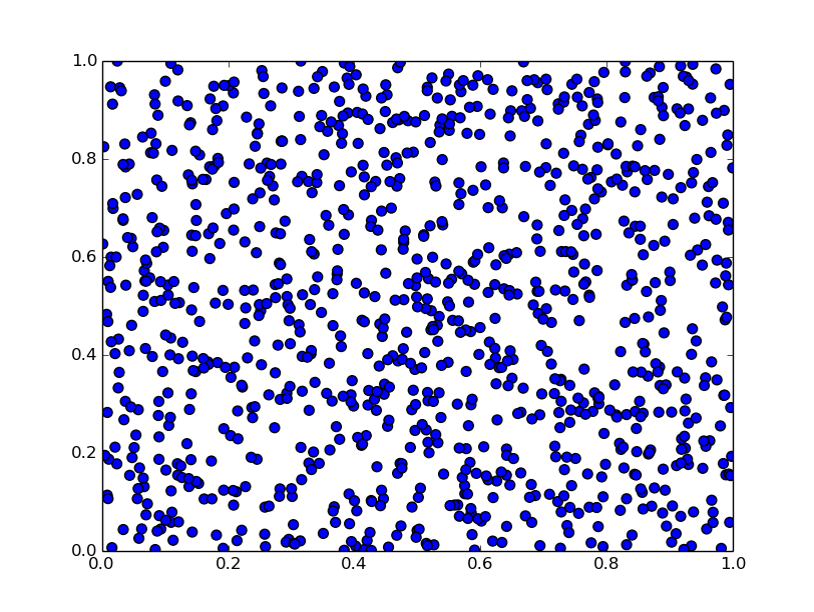}
    \subcaption{500 2D randomly generated sequences}
    \end{minipage}\hfill
    \begin{minipage}{.5\textwidth}
    \includegraphics[width=\textwidth]{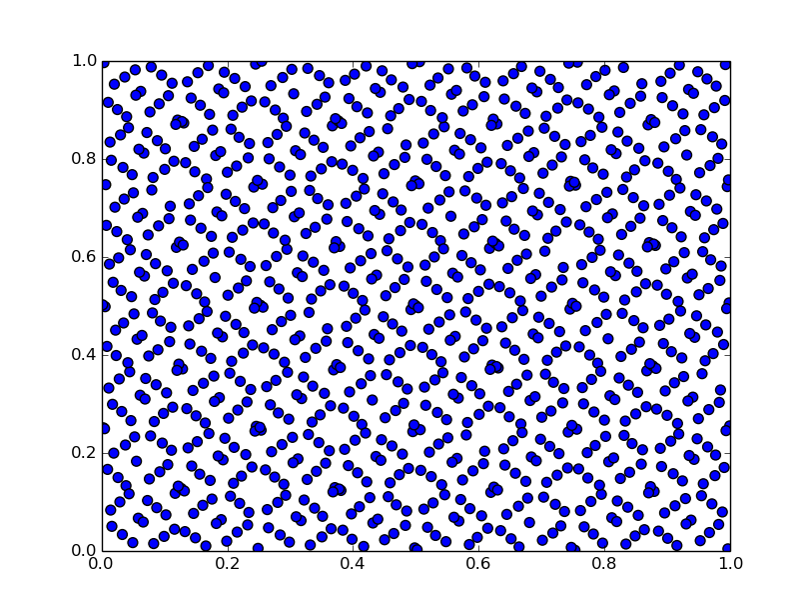}
    \subcaption{500 2D Sobol sequences}
    \end{minipage}
    \caption{Randomly generated sequences leave larger voids compared to Sobol sequences in the same configuration space.}
    \label{fig:seq}
\end{figure}

Each sample is a unique configuration of the robot's arm in its C-space which can belong to any one of the three categories: 
\begin{itemize}
\item self-collision - a link of the arm is in collision with another
\item $C_{obs}$ - arm is in collision with some object in the environment
\item $C_{free}$ - not in any collision at all
\end{itemize}

Obviously, it can happen that a sample is both in self-collision and in $C_{obs}$. In those cases, we consider the sample to be in self-collision.

Collision checker module is called on the sample to determine its collision state. If it in self-collision it is discarded, otherwise included in the training set. We maintain an equal number of $C_{free}$ samples and $C_{obs}$ samples in the training set. In this way, we generate a training set of $N$ samples comprising of $N/2$ $C_{free}$ and $N/2$ $C_{obs}$ samples. Moreover, for each of the $C_{obs}$ samples, a collision report is obtained from the collision checker module which contains information regarding the 3D position and orientation of all the in-collision points of arm. Pseudocode for generating training set having $N$ samples is shown below:
\begin{lstlisting}
generate_training_set(N_train):
	Cfree_samples  = []
	Cobs_samples = []
	collision_reports = []
	seed = 1
	while Cobs_sample_count < N_train/2
  		sobol_sequence = generate_sobol_sequence(seed)
  		seed = seed + 1  
  		sample = transform_to_Cspace(sobol_sequence)
  		state = collision_check(sample,report)
  		if state == Cfree
    			add sample to Cfree_samples
  		elif state == Cobs
    			add sample to Cobs_samples
    			add report to collision_reports
	Cfree_samples = C_free_samples[1:N_train/2] 
	return Cfree_samples, Cobs_samples, collision_reports     
\end{lstlisting}

The belief of a query configuration \textbf{q'} to be collision-free is evaluated as the weighted average of the collision state (-1 for $C_{free}$ and 1 for $C_{obs}$ sample) of some of its neighbouring configurations from the training set. 

\begin{equation*}
belief(\mathbf{q'}) = \frac{\Sigma_{q \in P}w_{q}S(q)}{\Sigma_{q \in P}w_{q}}
\end{equation*}
where, $P$ is the set of neighbouring configurations of query $\mathbf{q'}$, $w_{q}$ and $S(q)$ are the weight and true collision state of neighbour $\mathbf{q}$ respectively.  
Here, there are two parameters in estimating belief are set $P$, and weights $w_{q}$ $\forall q\in P$. 

The neighbouring configurations are found out using various methods as discussed in the subsequent subsections. The weight of a neighbouring configuration \textbf{q} is a measure of its importance in determining the collision probability of \textbf{q'}. Closer is \textbf{q} from \textbf{q'}, higher is the probability of \textbf{q'} sharing the same collision state as \textbf{q} as per the assumption of continuous ambient space, hence higher is the weight of \textbf{q}.

\begin{figure}[ht]
    \begin{minipage}{.5\textwidth}
        \includegraphics[width=\textwidth]{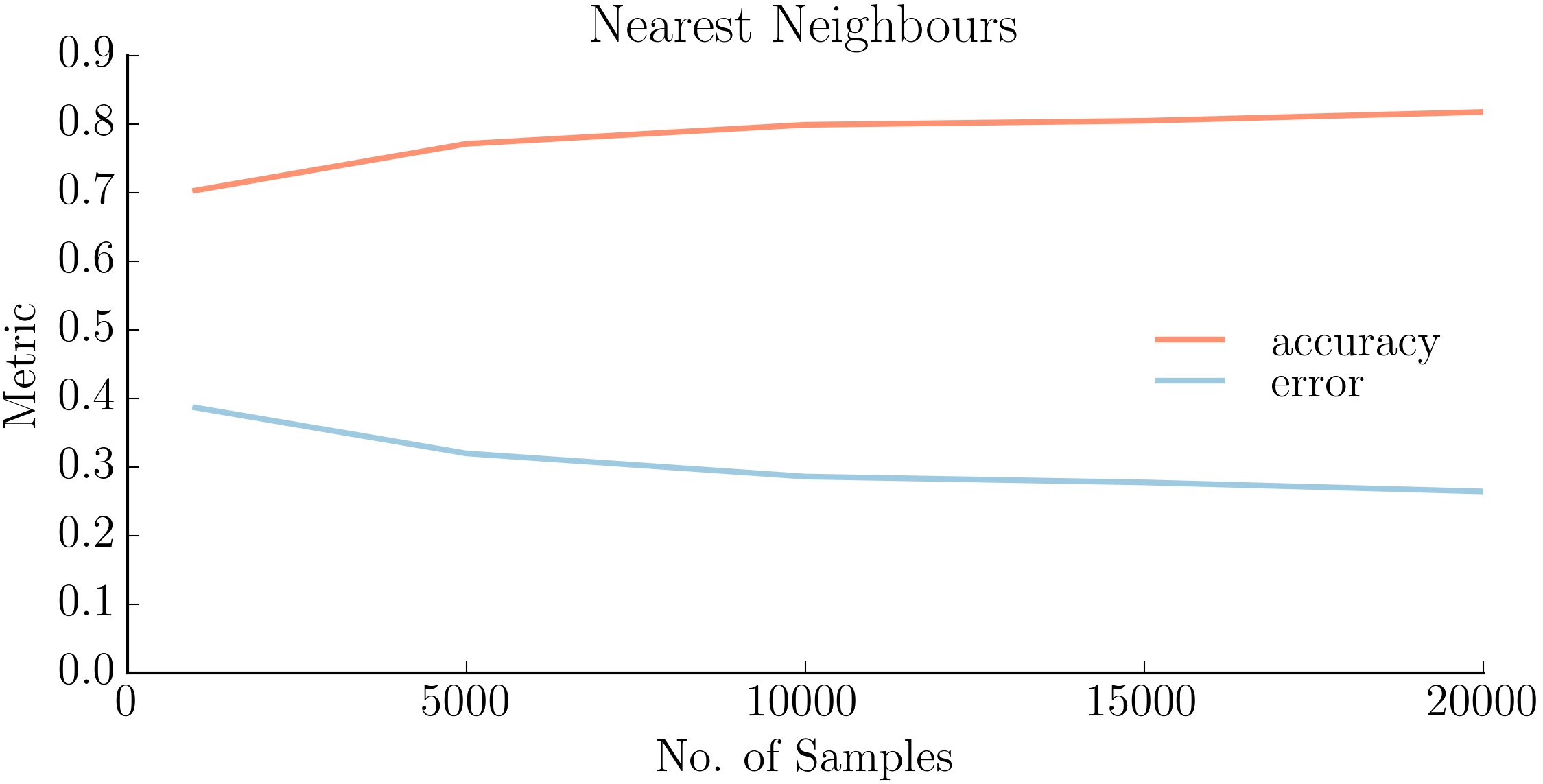}
        \subcaption{Fixed k}
        \label{fig:nn_k}
    \end{minipage}
    \begin{minipage}{.5\textwidth}
        \includegraphics[width=\textwidth]{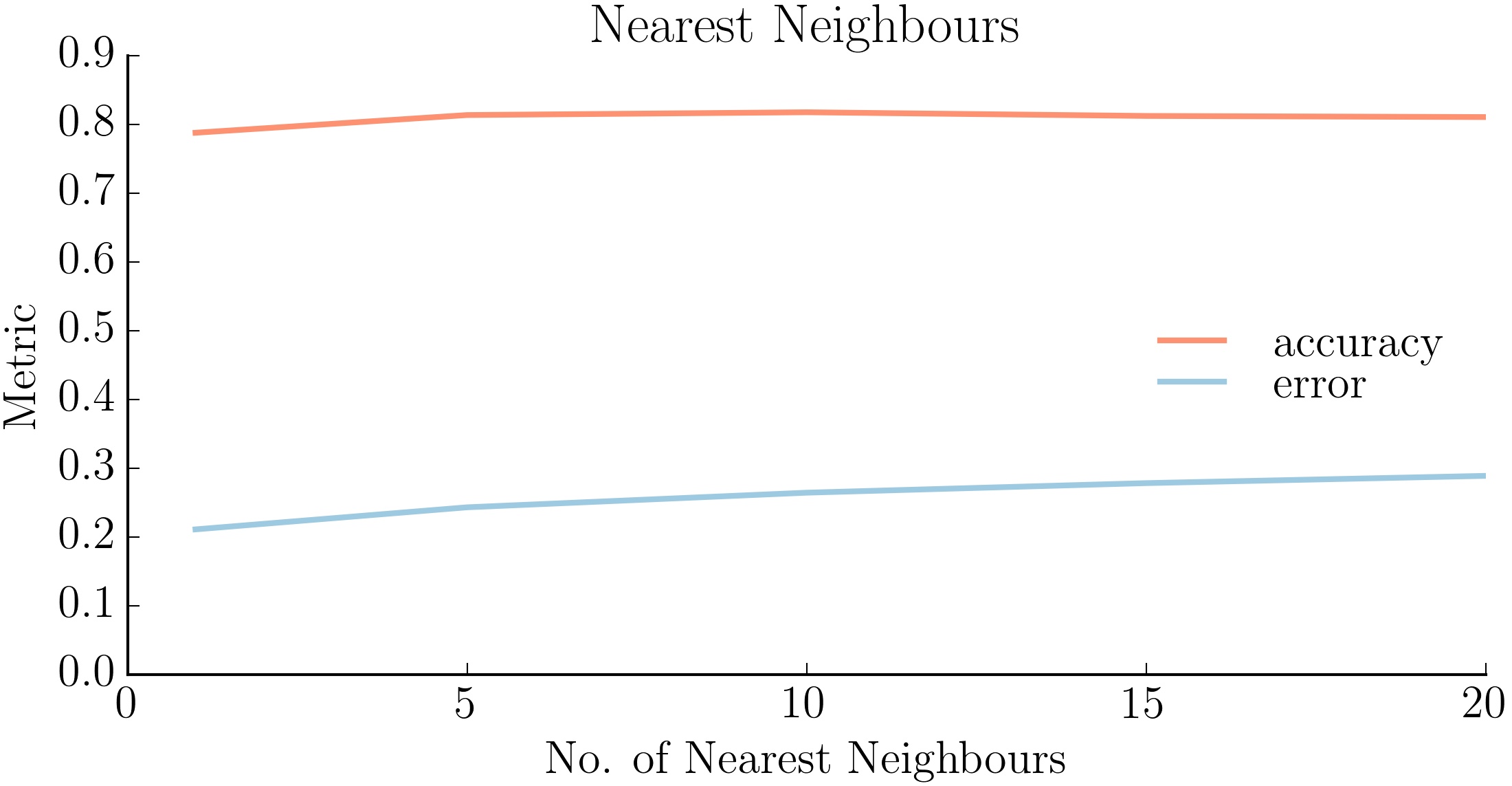}
        \subcaption{Fixed N}
        \label{fig:nn_n}
    \end{minipage}
\caption{(a) shows the performance of NN method as a function of $N$ for $k=10$. In (b), NN method is evaluated for different values of $k$ for $N=20000$.}
\label{fig:nn}
\end{figure}

\begin{figure}[t]
    \begin{minipage}{.5\textwidth}
    \includegraphics[width=\textwidth]{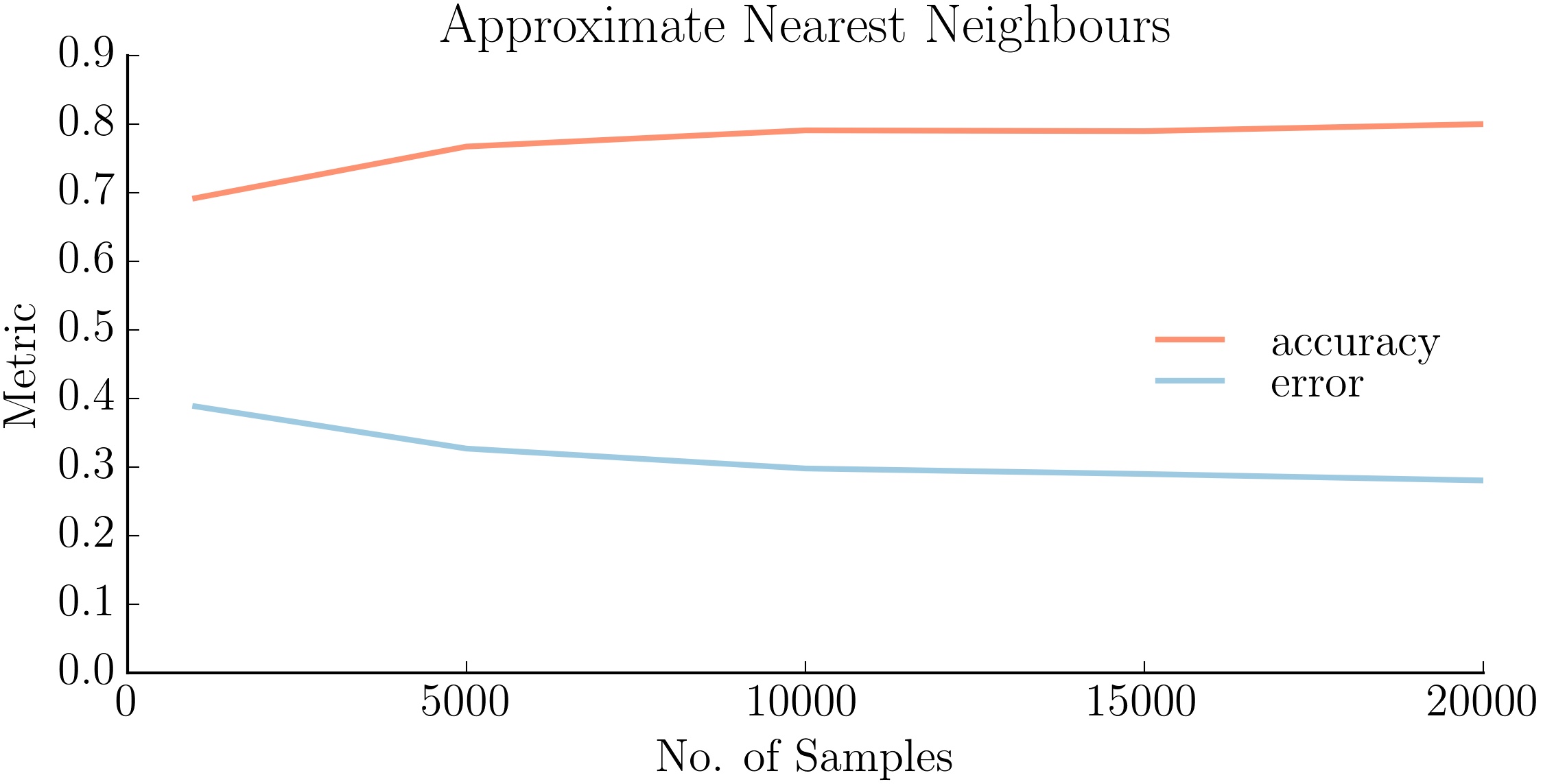}
    \subcaption{Fixed k}
    \label{fig:ann_k}
    \end{minipage}
    \begin{minipage}{.5\textwidth}
    \includegraphics[width=\textwidth]{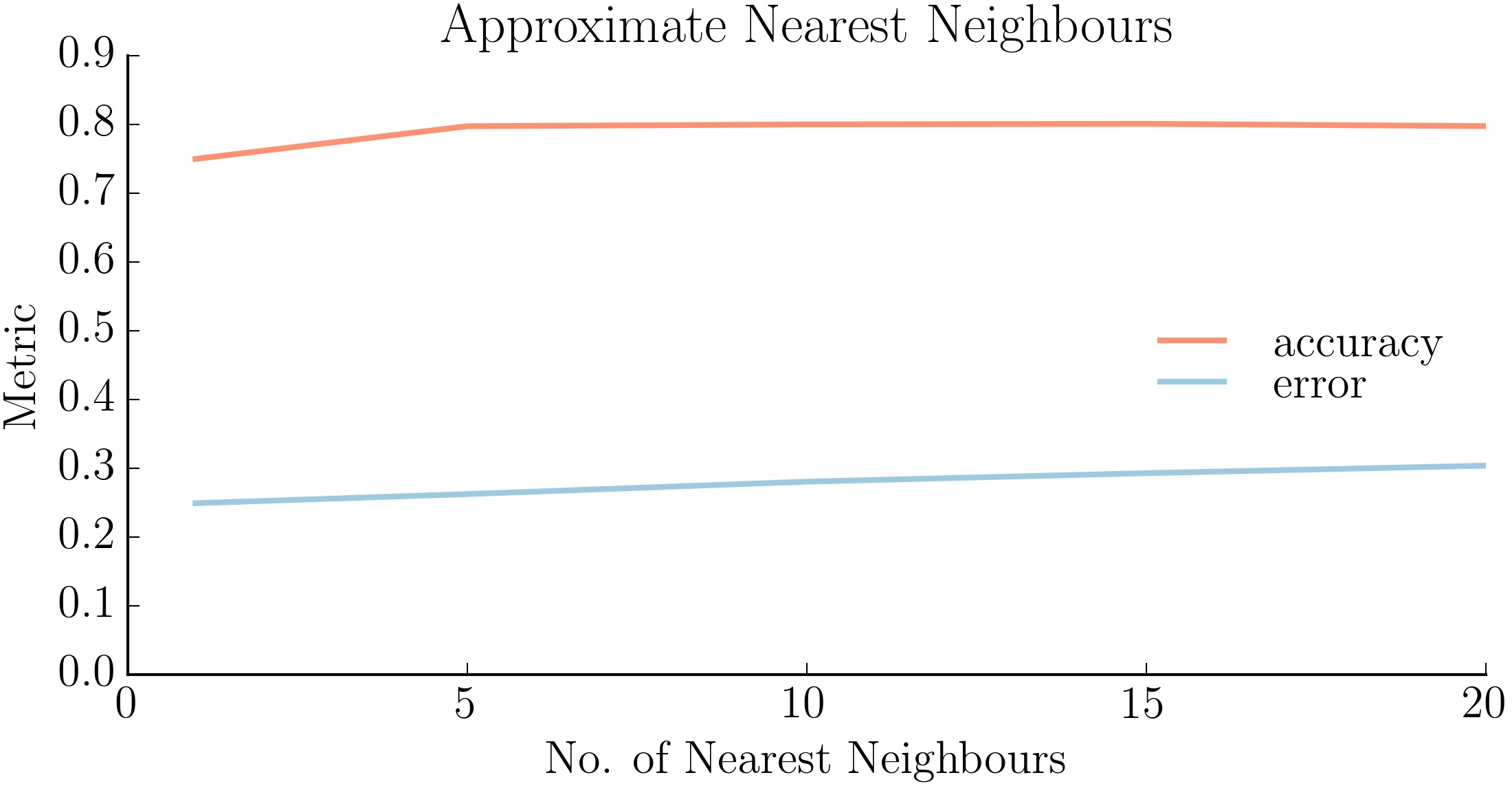}
    \subcaption{Fixed N}
    \label{fig:ann_n}
    \end{minipage}
\caption{In (a), the performance of ANN method as a function of $N$ is shown for a fixed $k=10$. In (b), ANN method is evaluated for different values of $k$ for $N=20000$.}
\label{fig:ann}
\end{figure}

\subsection{Exact Nearest Neighbours}
The first method used to find the neighbouring configurations is the exact nearest neighbours or simply the nearest neighbour (NN) method ~\cite{altman1992introduction}. $k$ nearest neighbours of the query configuration $\mathbf{q}$ collectively  evaluate its collision probability or the belief of being collision-free. The weight of configuration $\mathbf{q'}$  is taken to be the reciprocal of the euclidean distance between $\mathbf{q'}$  and $\mathbf{q}$, i.e, $$w_{q} = \frac{1}{\|\mathbf{q}-\mathbf{q'}\|}$$ 
This weight metric is consistent with the assumption of continuous ambient space. Larger the value of $\frac{1}{\|\mathbf{q}-\mathbf{q'}\|}$, closer are \textbf{q'} and \textbf{q}, hence higher should be the weight. Figure \ref{fig:nn_k} shows the performance of NN method as a function of $k$ for a fixed value of $N$ (size of training set). We also varied $N$ keeping $k$ constant and the corresponding results are shown in Figure \ref{fig:nn_n}.

\subsection{Approximate Nearest Neighbours}
Although NN method gives the $k$ closest training samples to the query but is computationally expensive both from the time and memory consumption point of view. In such circumstances, it may be a good idea to settle for a 'good guess' of nearest neighbours in the exchange of reduced computational load. Approximate Nearest Neighbours ~\cite{chakrabarti2004optimal} (ANN) method fulfills the requirements. Although, it doesn't guarantee to return the closest neighbours but is computationally faster and consumes less memory than the NN method. After determining the required number of neighbours, their weights are computed in the exact same manner as in the case of NN method. Figure \ref{fig:ann_k} shows the performance of ANN method as a function of $k$ for a fixed $N$. Also, the results of varying $N$ keeping $k$ constant are shown in Figure \ref{fig:ann_n}.

\subsection{Fixed Radius Near Neighbours}
Both the NN and ANN methods find a pre-determined number of neighbours of a query configuration \textbf{q'}. However, it may be useful to relax this constraint on the quantity of neighbours and instead determine all the neighbours residing inside a 7-dimensional sphere of radius $r$ centred at \textbf{q'}. In this case, we computed the weight $w_{q}$ of configuration \textbf{q} using either of the two kernels: 
\begin{itemize}
    \item Gaussian kernel: It is a second-order non-parametric kernel. However, it is not compact as its support runs over the entire space. $$ w_{q}= exp(-\frac{\|\mathbf{q}-\mathbf{q'}\|^2}{\sigma^2}) $$ where $\sigma$ is the variance of the training set.
    \item Epanechnikov kernel: It is a smooth non-parametric kernel. Among all the kernels it minimizes asymptotic mean integrated square error and is therefore optimal.
$$ w_{q}= \begin{cases}
\frac{3}{4} (1- \frac{\|\mathbf{q}-\mathbf{q'}\|^2}{r^2}) \text{ if } \|\mathbf{q}-\mathbf{q'}\|^2 < {r^2} \\
0 \text{ otherwise}
\end{cases} $$ 
\end{itemize}

\begin{figure}[ht]
    \begin{minipage}{.5\textwidth}
        \includegraphics[width=\textwidth]{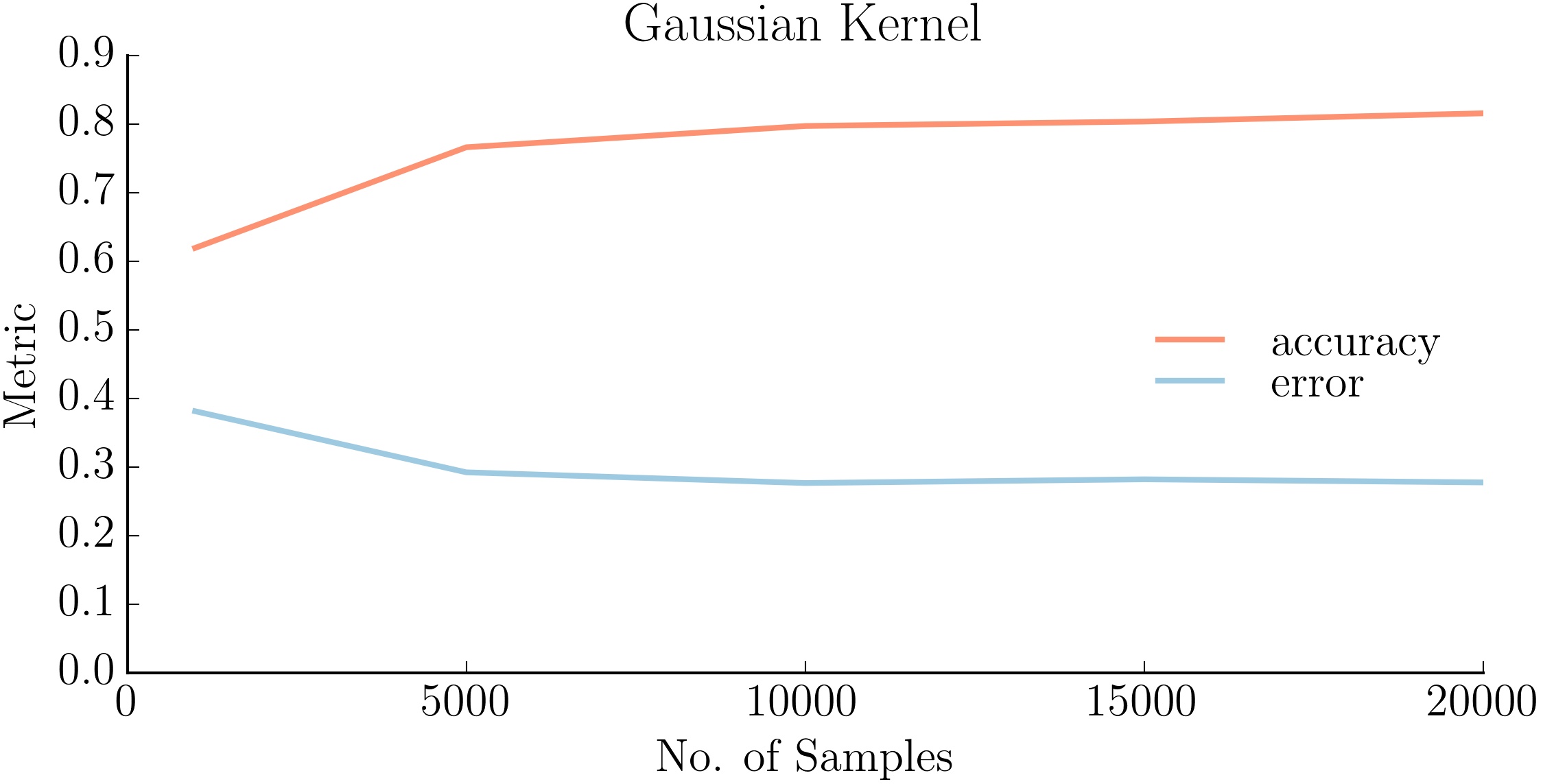}
        \subcaption{Fixed k}
        \label{fig:gk_k}
    \end{minipage}
    \begin{minipage}{.5\textwidth}
        \includegraphics[width=\textwidth]{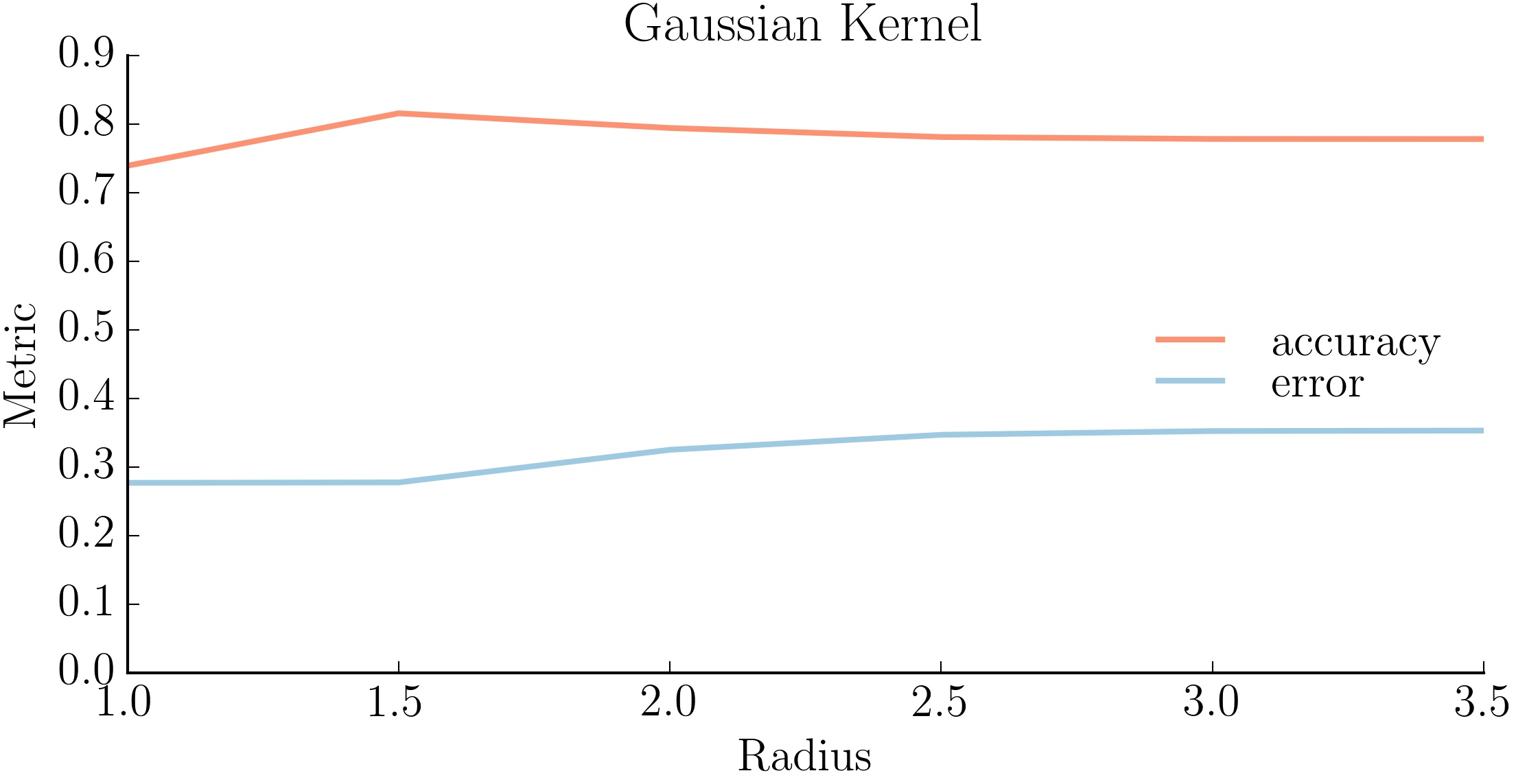}
        \subcaption{Fixed N}
        \label{fig:gk_n}
    \end{minipage}
    \caption{(a) shows the performance of a model using Gaussian kernel as a function of $N$ for $r=1.5$. In (b), the same model is evaluated for different values of $r$ for $N=20000$.}
    \label{fig:gk}
\end{figure}

\begin{figure}[ht]
    \begin{minipage}{.5\textwidth}
        \includegraphics[width=\textwidth]{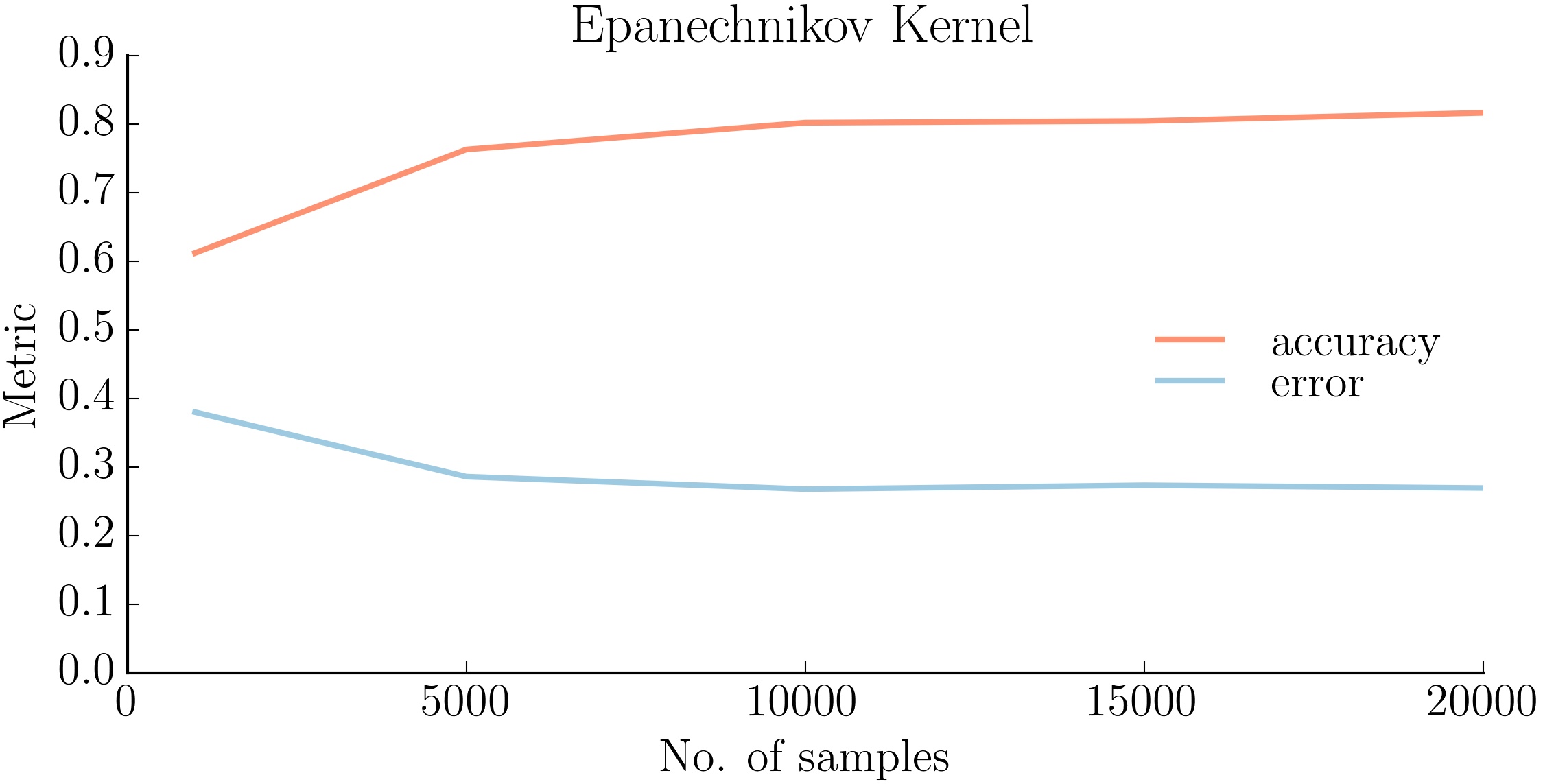}
        \subcaption{Fixed k}
        \label{fig:ep_k}
    \end{minipage}
    \begin{minipage}{.5\textwidth}
        \includegraphics[width=\textwidth]{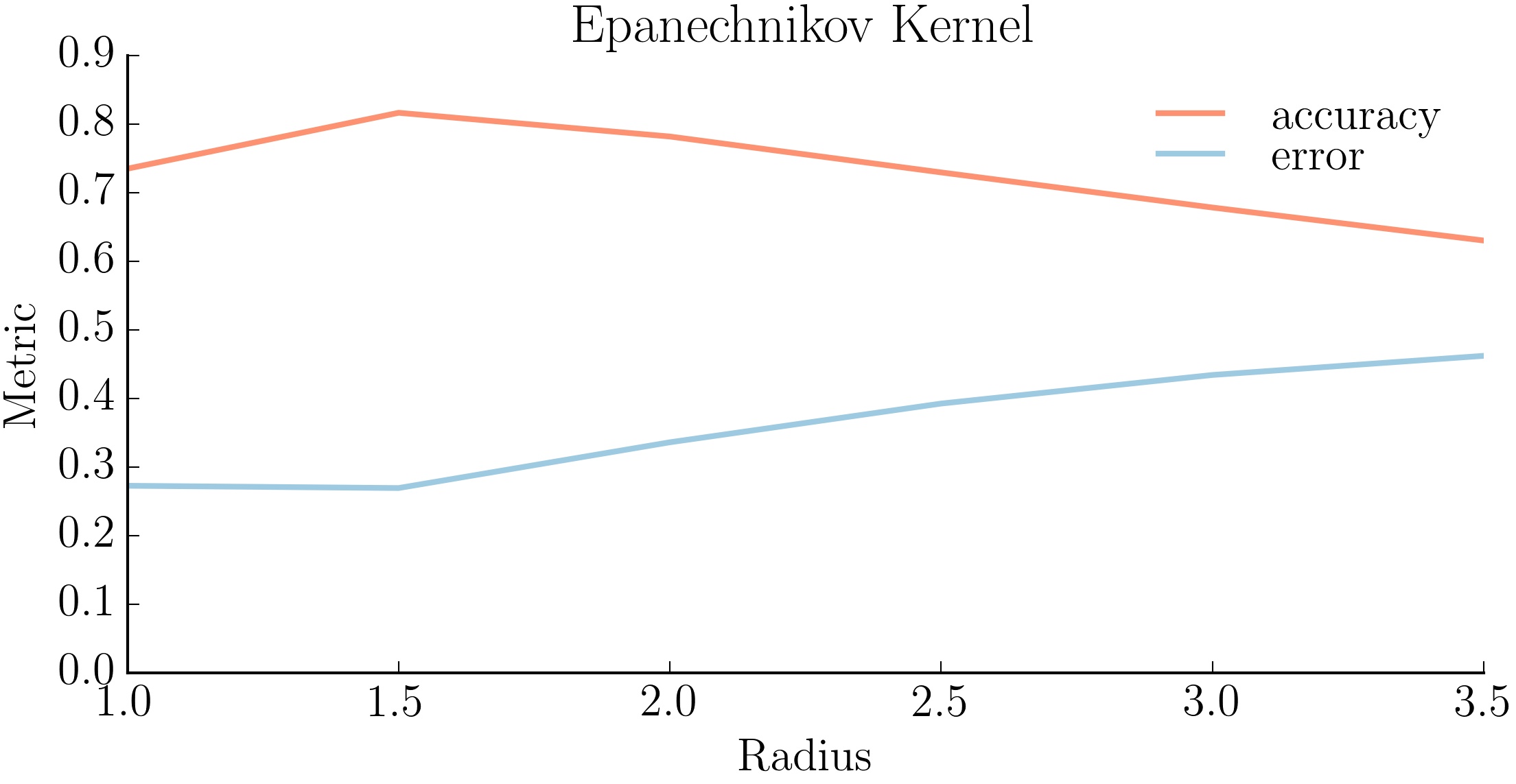}
        \subcaption{Fixed N}
        \label{fig:ep_n}
    \end{minipage}
    \caption{(a) shows the performance of a model using Epanechnikov kernel as a function of $N$ for $r=1.5$. In (b), the same model is evaluated for different values of $r$ for $N=20000$.}
    \label{fig:ep}
\end{figure}

The results of using Gaussian and Epanechnikov kernel ~\cite{epanechnikov1969non} are shown in Figure \ref{fig:gk} and \ref{fig:ep} respectively. 

Each of the three methods used to compute $w_{q}$ is a function of the similarity (or closeness) of neighbour \textbf{q} to query \textbf{q'}. We refer to this quantity as the similarity estimate between configurations \textbf{q} and \textbf{q'} denoted by $sim(q,q')$. Table \ref{tab:dist} lists the three methods used to estimate $w_{q}$.
\begin{table}[]
    \centering
        \begin{tabular}{|c|c|}
        \hline Distance metric & $w_{q}$ \\ \hline
        Inverse & $\frac{1}{sim(\mathbf{q},\mathbf{q'})}$ \\ \hline
        Gaussian & $ exp(-\frac{sim(\mathbf{q},\mathbf{q'})^2}{\sigma^2}) $\\ \hline
        Epanechnikov & $\frac{3}{4} (1- \frac{sim(\mathbf{q},\mathbf{q'})^2}{r^2})$ \\ \hline
        \end{tabular}
    \caption{Distance metrics and their corresponding weight formula}
    \label{tab:dist}
\end{table}

\subsection{Importance weights}

\begin{figure}[ht]
    \centering
    \includegraphics[width=\textwidth]{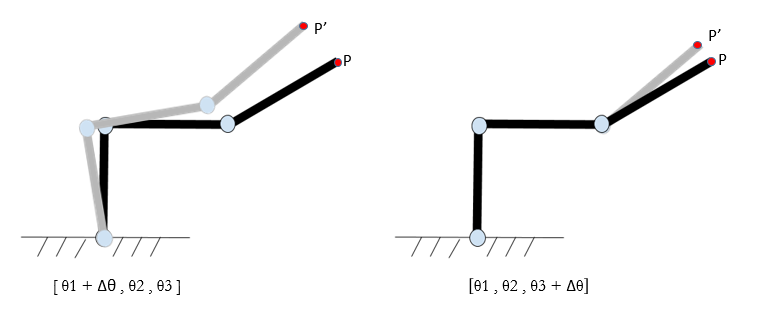}
    \caption{The black links show the initial configuration of 3R manipulator whereas the grey ones show the neighbouring ones. With respect to the original configuration, the grey link on the right is a closer neighbour than is the one on the left.}
    \label{fig:mani}
\end{figure}

All the three weight metrics discussed so far- reciprocal of euclidean distance, Gaussian and Epanechnikov kernels- compute $\|\mathbf{q}-\mathbf{q'}\|$ giving equal importance to all the dimensions of the C-space. Physically, it means that all the joints of the arm are assigned equal weight in estimating the C-space belief. In other words, what matters is the absolute difference between \textbf{q'} and \textbf{q} and not the individual difference in each joint values. But, intuitively not all the joints should be equally important. For any manipulator, the base joint have more influence on a configuration than the wrist joints as a small perturbation in the base joint results in the motion of all the links while that in wrist joint (closer to the end-effector) affect only the wrist ones. This whole idea is illustrated in case of a planar 3R manipulator in Figure \ref{fig:mani}. 

Let, $\mathbf{q'}=[\theta_{1}; \theta_{2}; \theta_{3}]$ be the query configuration whose belief of being collision-free is to be estimated. We take two neighbouring configurations $\mathbf{q_{1}} = [\theta_{1}+\Delta \theta, \theta_{2}, \theta_{3}]$ and $\mathbf{q_{2}}=[\theta_{1}, \theta_{2}, \theta_{3}+\Delta \theta]$. Clearly, the weight of the two configurations $\mathbf{q_{2}}$ and $\mathbf{q_{2}}$ are equal as $\|\mathbf{q'}-\mathbf{q_{1}}\| = \|\mathbf{q'}-\mathbf{q_{2}}\|$. However, as we can observe from the figure itself, $\mathbf{q_{2}}$ is a closer neighbour of $\mathbf{q'}$ than is $\mathbf{q_{1}}$, hence $\mathbf{q_{2}}$ should have a comparatively higher weight than $\mathbf{q_{1}}$ in estimating the collision probability of $\mathbf{q'}$. This ambiguity can be resolved by assigning importance weights to each joint of the robot's arm and then computing weighted L2 norm rather than the simple L2 norm. We will refer to these weights as 'importance weight' of joint in order to avoid any confusion with the weight of the training samples.

The importance weight of a joint is a measure of its dominance on the arm configuration. A joint which when actuated by a small amount results in a large deviation from the current configuration of the arm will have higher importance weight than others. Intuitively, shoulder joints should have higher weights than wrist joints. From the very definition, it is clear that the importance weights depend on the configuration itself. So, rather than allotting a constant weight to each joint, we assigned weight to each joint of a configuration depending on its ability to disturb the original state of the arm. Moreover, the interpretation and hence the calculation of importance weights of a $C_{free}$ sample is different from that of a $C_{obs}$ sample as explained below. We denote the vector of the importance weight of configuration $\mathbf{q}$ as $\mathbf{\Omega_{q}}$. The importance weight of $j^{th}$ joint of $C$ is given by the $j^{th}$ element of $\Omega_{q}[j]$.

For $C_{free}$ samples, the deviation from the original configuration is measured as the deflection in the position of the end-effector when a joint is perturbed by 0.01 radian.  the initial position of end-effector be $\mathbf{v}$. $\mathbf{\Omega_{q}}$ can be computed as follows:
\begin{description}
\item[1.] Increase $q[j]$ by 0.01 radians keeping all other joint angles fixed
\item[2.] Determine the new position $\mathbf{v'_{j}}$ of the end-effector
\item[3.] Compute $ s_{qj} = \|\mathbf{v'_{j}} - \mathbf{v}\|$
\end{description}

Construct a vector $\mathbf{s_{q}}$ whose $j^{th}$ element is $s_{qj}$ as computed above. $\mathbf{\Omega_{q}}$ is given by the following formula:
$$ \mathbf{\Omega_{q}} = \frac{\mathbf{s_{q}}}{\|\mathbf{s_{q}}\|} $$
 
In case of $C_{obs}$ samples, the calculation of $\mathbf{\Omega_{q}}$ is a bit more involved. Since some link of the arm is in collision with some object(s) in the environment, the ability of a joint to perturb the configuration from its original state is not the same as in the previous case. Depending on the link in-collision, some joints can alter the collision state while the rest can't. In order to calculate $\mathbf{\Omega_{q}}$ for $C_{obs}$ sample $\mathbf{q}$, we need to compute two things beforehand with the help of collision report: 

\begin{itemize}
    \item the link in-collision with the environment. Although there can be multiple such links, we have considered only the first link, i.e., the one closest to the shoulder of HERB. Let the index of this link be $l$.
    \item the centre of mass P of all the in-collision meshes of link $l$. Lets call the vector to be $\mathbf{v_{P}}$ 
\end{itemize}

According to our convention, both joint index and link index start from 0 with link 0 being the fixed link and joint $j$ connects link $j$ and link $j+1$. Note that if link $l$ of sample $\mathbf{q}$ is in collision, then actuating any joint between link $l$ and the end-effector will not affect the link $l$ hence the collision state of the arm. In other words, whatever be the values of joint $l, l+1, .. , 6$, the arm will always be in collision. This suggests that the difference in values of these joints between query $\mathbf{q'}$ and $\mathbf{q}$ is immaterial in the computation of the weight of $\mathbf{q}$. Hence, the importance weight of these joints is assigned to be zero. However, the remaining joints can affect the state of configuration $\mathbf{q'}$. The importance weight of joint $j$ ($0\leq j\leq l-1 $) is computed in a similar manner to that in $C_{free}$ configurations. Only this time the role of end-effector is supplanted by the point P. $\mathbf{\Omega_{q}}$ can be computed as follows:
\begin{description}
\item[1.] Increase $q[j]$ by 0.01 radians keeping all other joint angles fixed
\item[2.] Determine the new position $\mathbf{v'_{P}}$ of the point P
\item[3.] Compute $ s_{qj} = \|\mathbf{v'_{P}} - \mathbf{v_{P}}\|$
\end{description}
For $j \geq l$ , $ s_{qj} = 0$. We construct a 7-dimensional vector $\mathbf{s_{q}}$ whose $j^{th}$ element is $s_{qj}$ as computed above. Finally, 
$$ \mathbf{\Omega_{q}} = \frac{\mathbf{s_{q}}}{\|\mathbf{s_{q}}\|} $$

For sample $\mathbf{q}$, we construct a 7X7 diagonal matrix $\mathbf{W_{q}}$ whose $i^{th}$ diagonal element is ${\Omega_{q}[i]}$. The weighted L2 norm can thus be computed as:
$$\|\mathbf{q}-\mathbf{q'}\|_{w}^{2} = (\mathbf{q}-\mathbf{q'})^{T}\mathbf{W_{q}}(\mathbf{q}-\mathbf{q'})$$
 
So far, we have mentioned two measures of estimating similarity between two configurations $\mathbf{q}$ and $\mathbf{q'}$ - L2 norm and weighted L2 norm. This similarity estimate is then used in the calculation of weight of a training sample. In addition, we have also used scale-invariant mahalanobis distance~\cite{mahalanobis1936generalized} as the similarity measure. In sum, we have used four different measures for estimating similarity between configurations $\mathbf{q}$ and $\mathbf{q'}$ as summarised below:  

\begin{center}
\begin{tabular}{|c|c|}
\hline 
Measure & $sim(\mathbf{q},\mathbf{q'})^{2}$\\ \hline
Euclidean & $(\mathbf{q}-\mathbf{q'})^T(\mathbf{q}-\mathbf{q'})$ \\ \hline
Weighted Euclidean & $(\mathbf{q}-\mathbf{q'})^T\mathbf{W_{q}}(\mathbf{q}-\mathbf{q'}) $  \\ \hline
Mahalanobis & $ (\mathbf{q}-\mathbf{q'})^T\mathbf{\Sigma}^{-1}(\mathbf{q}-\mathbf{q'}) $ \\ \hline
Weighted Mahalanobis & $(\mathbf{q}-\mathbf{q'})^T\mathbf{W_{q}}\mathbf{\Sigma}^{-1}(\mathbf{q}-\mathbf{q'}) $ \\ \hline
\end{tabular}
\end{center}
where $\mathbf{\Sigma}$ is the covariance matrix of the training set. 

\subsection{Topological method}
All the methods discussed so far to compute the C-space belief  are applicable in the 7D C-space of the robot and does not capture any higher order topological structure of the environment. Exploiting the embedded structure has proved to be of significant importance in motion planning. The methods that seek to determine and exploit such structures and  the topology of the space are called topological methods. They do not rely on the exact metric information and hence are robust to errors. 

However, the computational complexity of such methods increase exponentially with the dimensionality of the space. Hence, practically these methods can't be applied in a high dimensional space like the 7-D C-space. We solved the high dimensionality problem by projecting the 7-D C-space to a 4-D space referred as the 'topological space'. For any vector $\mathbf{v}$ in the C-space, its projection $\mathbf{v^{p}}$ in the topological space is given by the following rule: 
$$ \mathbf{\overline{v}} = \mathbf{v}[1:4] $$
where $\mathbf{v}[1:4]$ means the first four elements of vector $\mathbf{v}$. 
This projected space is constructed by removing the last 3 axes of the C-space. In other words, it is the state space of the first four arm joints, i.e., shoulder and elbow joints. The motivation behind this projection rule is embedded in the mechanical design of the arm itself. HERB's arm has 2 shoulder, 2 elbow and 3 wrist joints with the wrist links being much smaller than the other links. Since the latter joints can actuate only the wrist links, they have little influence on the arm configuration as compared to the shoulder and elbow joints together. Hence neglecting them is a fairly reasonable step and does not result in much significant information loss.

\begin{figure}[ht]
\centering
\includegraphics[scale=.6]{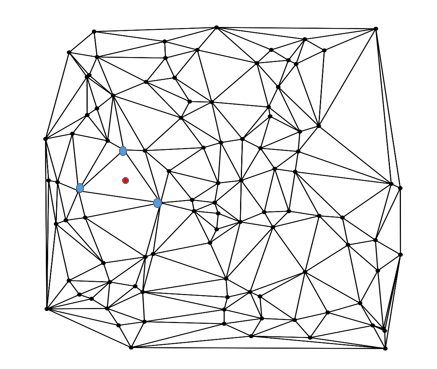}
\caption{Delaunay triangulation of a set of 2D points. The neighbours of the configuration shown by red dot are the three vertices shown with blue dots.}
\end{figure}

One of the ways to capture the topological structure of a set of 2D points is Delaunay triangulation. Triangulation of a set of 2D points results in the formation of triangles such that no point lies inside the circumcircle of any triangle. Tessellation in n-dimensional space is a general form of triangulation in 2-D space. Tessellation forms simplices such that no point lies inside the circum-hypersphere of any simplex. Simplex is the generalization of triangle or tetrahedron to any arbitrary dimension. A 2-simplex is the triangle and 3-simplex is tetrahedron. In general, a k-simplex is a k-dimensional polytope and the convex hull of its k+1 vertices. 
Some pre-processing is required:
\begin{description}
\item[1.] Project the entire training set to the topological space as per the projection rule 
\item[2.] Construct the Delaunay Tessellation of the projected training set \end{description}

Like in the previous methods, the belief of a configuration to be collision-free is the weighted average of the collision state of its neighbours. The steps to determine the neighbours of query $\mathbf{q'}$ are as follows:
\begin{description}
\item[1.] Find the projection $\mathbf{\overline{q'}}$ of query $\mathbf{q'}$ to the topological space
\item[2.] Determine the simplex which circumscribes $\mathbf{\overline{q'}}$. 
\end{description}

The neighbours of the  projected query configuration $\mathbf{\overline{q'}}$ are the vertices of the circumscribing simplex. Obviously, it is possible that $\mathbf{\overline{q'}}$ may not lie inside any simplex of the tessellation. In that case, we have assumed that the query is equally probable to be in $C_{free}$ or in $C_{obs}$ and hence its belief is 0.0 (remember in our case belief varies from -1.0 to 1.0).

In this method, the weight of a neighbour $\mathbf{\overline{q}}$ of query $\mathbf{\overline{q'}}$ is taken as the reciprocal of $sim(\mathbf{\overline{q}},\mathbf{\overline{q'}})$. In the case of weighted euclidean and weighted mahalanobis similarity measure, the importance weights used are also projected to the topological space using the projection rule. Moreover, the covariance matrix used in the mahalanobis and weighted mahanlanobis similarity measure is of the projected training. We call the belief of $\mathbf{q'}$ thus computed as $bel_{1}(\mathbf{q'})$. 

In order to make up for the lose of information due to projection, we compute a secondary belief $bel_{2}(\mathbf{q'})$. The process of estimating $bel_{2}(\mathbf{q'})$ is same as that of $bel_{1}(\mathbf{q'})$ except the projection step. All the configurations and their joints importance weights are projected to a 3D space by taking their last three joint values which is ignored in the topological space. Note that the neighbours of query $\mathbf{q}$ are estimated from the topological space only and the same neighbours are used to compute $bel_{2}(\mathbf{q'})$ also. The net belief of configuration $q$ to be collision-free is the weighted average of $bel_{1}(\mathbf{q'})$ and $bel_{2}(\mathbf{q'})$. Here, the weight of $bel_{1}(\mathbf{q'})$ is taken to be 100 times that of $bel_{2}(\mathbf{q'})$. 

\begin{figure}[t]
        \centering
        \begin{minipage}[b]{0.475\textwidth}
            \centering
            \includegraphics[width=\textwidth]{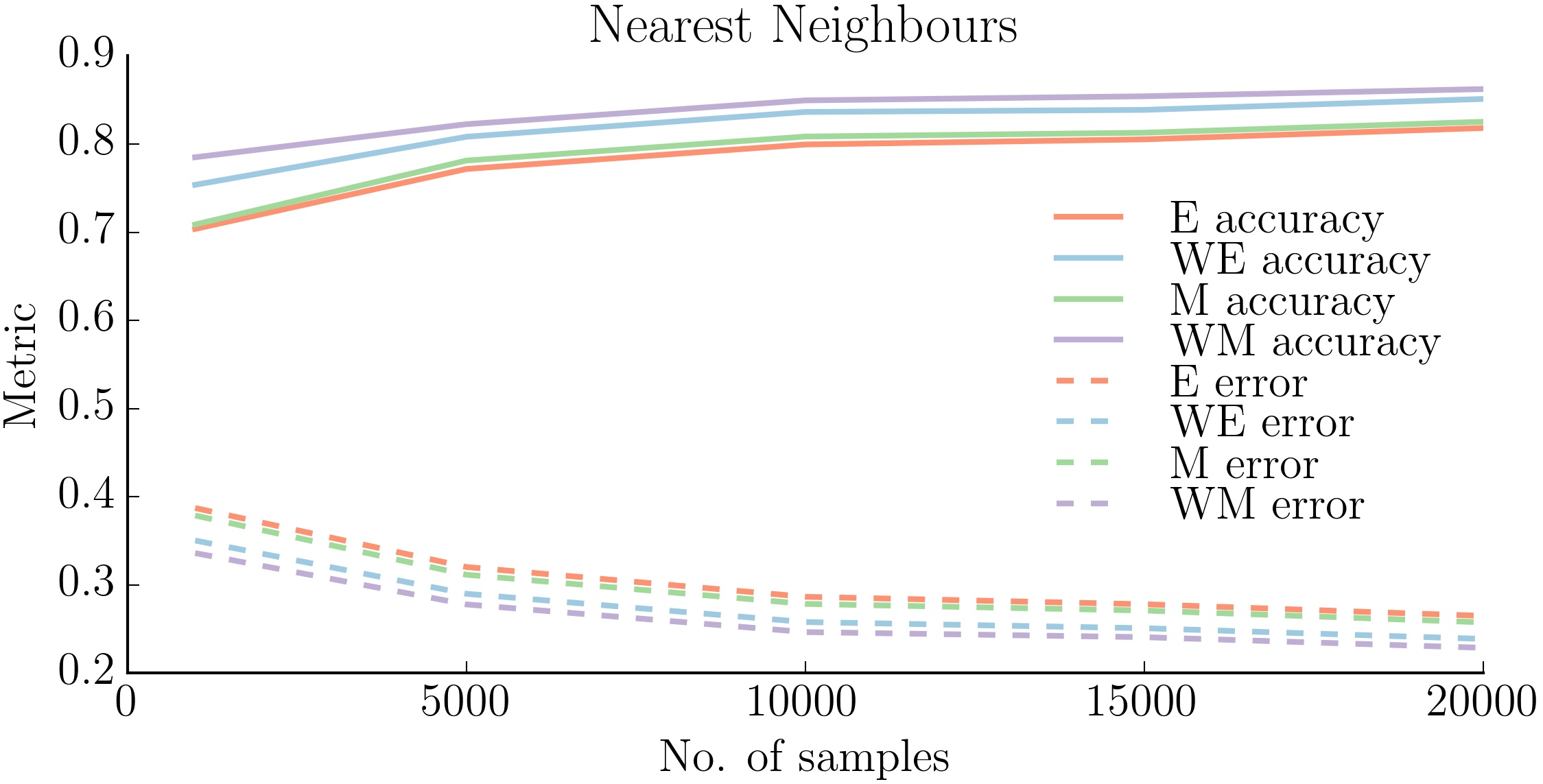}
            \subcaption{Nearest Neighbors}      
            \label{fig:nn_10}
            
        \end{minipage}
        \hfill
        \begin{minipage}[b]{0.475\textwidth}  
            \centering 
            \includegraphics[width=\textwidth]{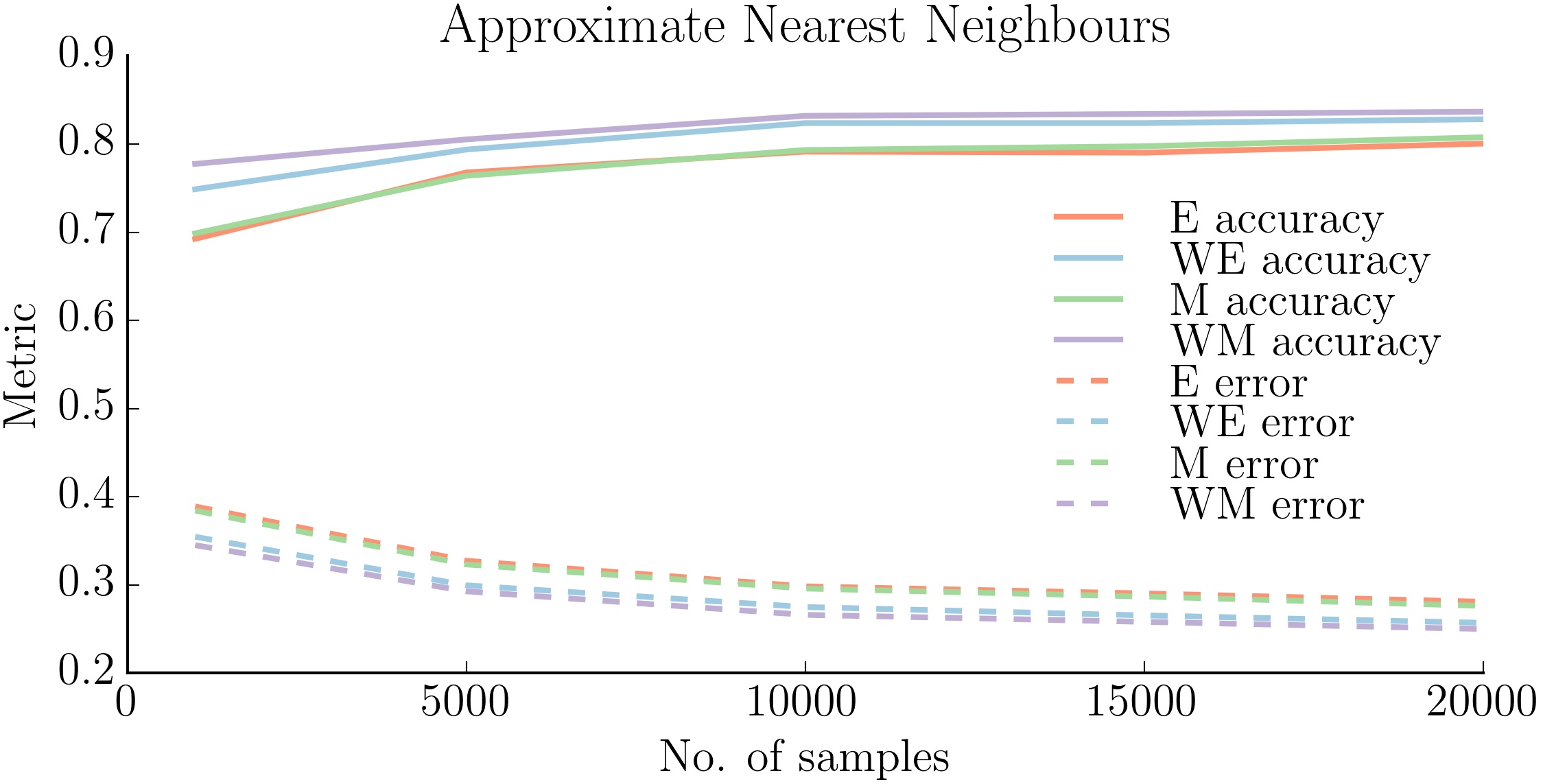}
            \subcaption{Approximate Nearest Neighbors}     
            \label{fig:ann_10}
            
        \end{minipage}
        \vskip\baselineskip
        \begin{minipage}[b]{0.475\textwidth}   
            \centering 
            \includegraphics[width=\textwidth]{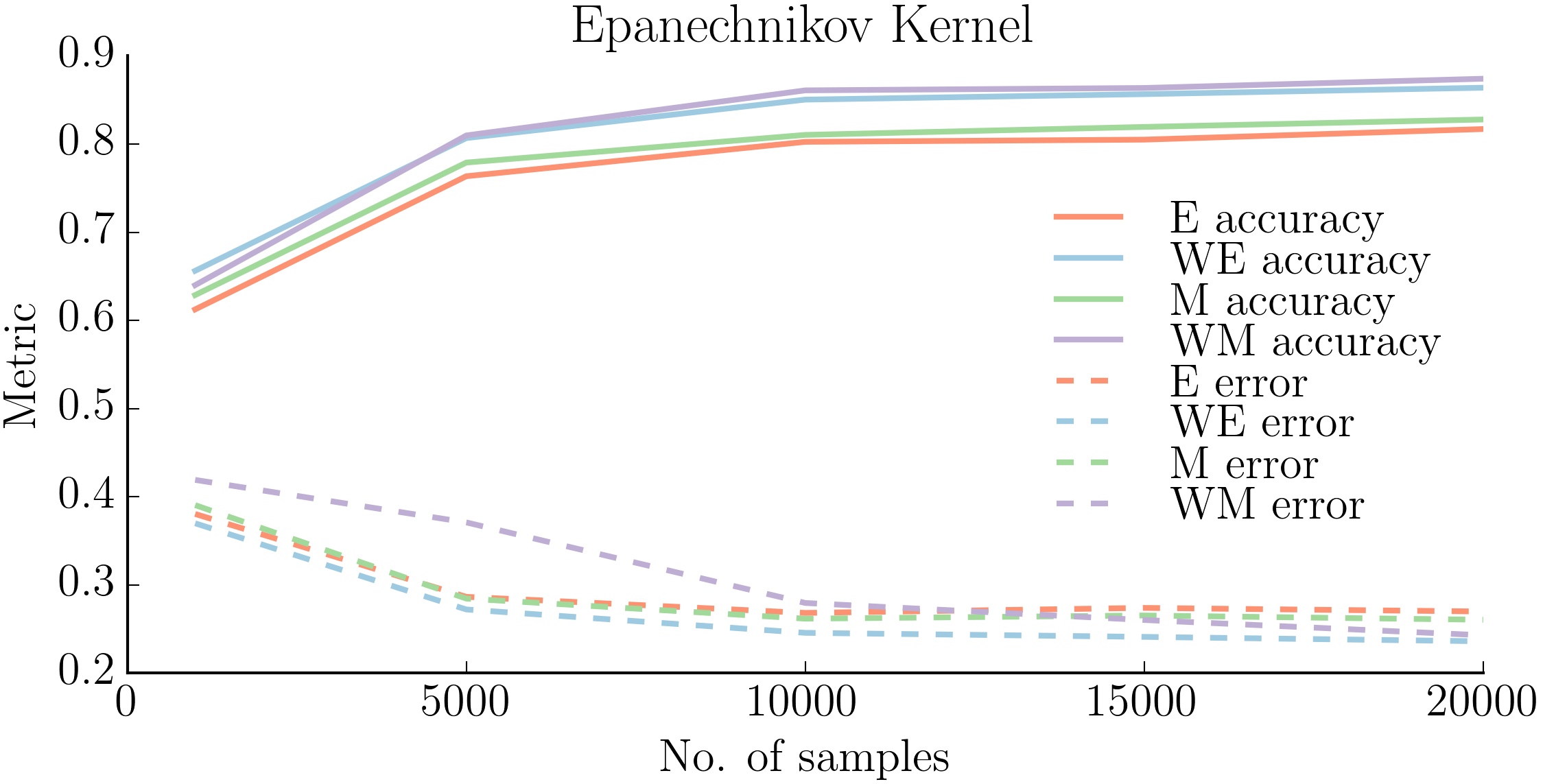}
            \subcaption{Epanechnikov Kernel}     
            \label{fig:ep_15}
            
        \end{minipage}
        \quad
        \begin{minipage}[b]{0.475\textwidth}   
            \centering 
            \includegraphics[width=\textwidth]{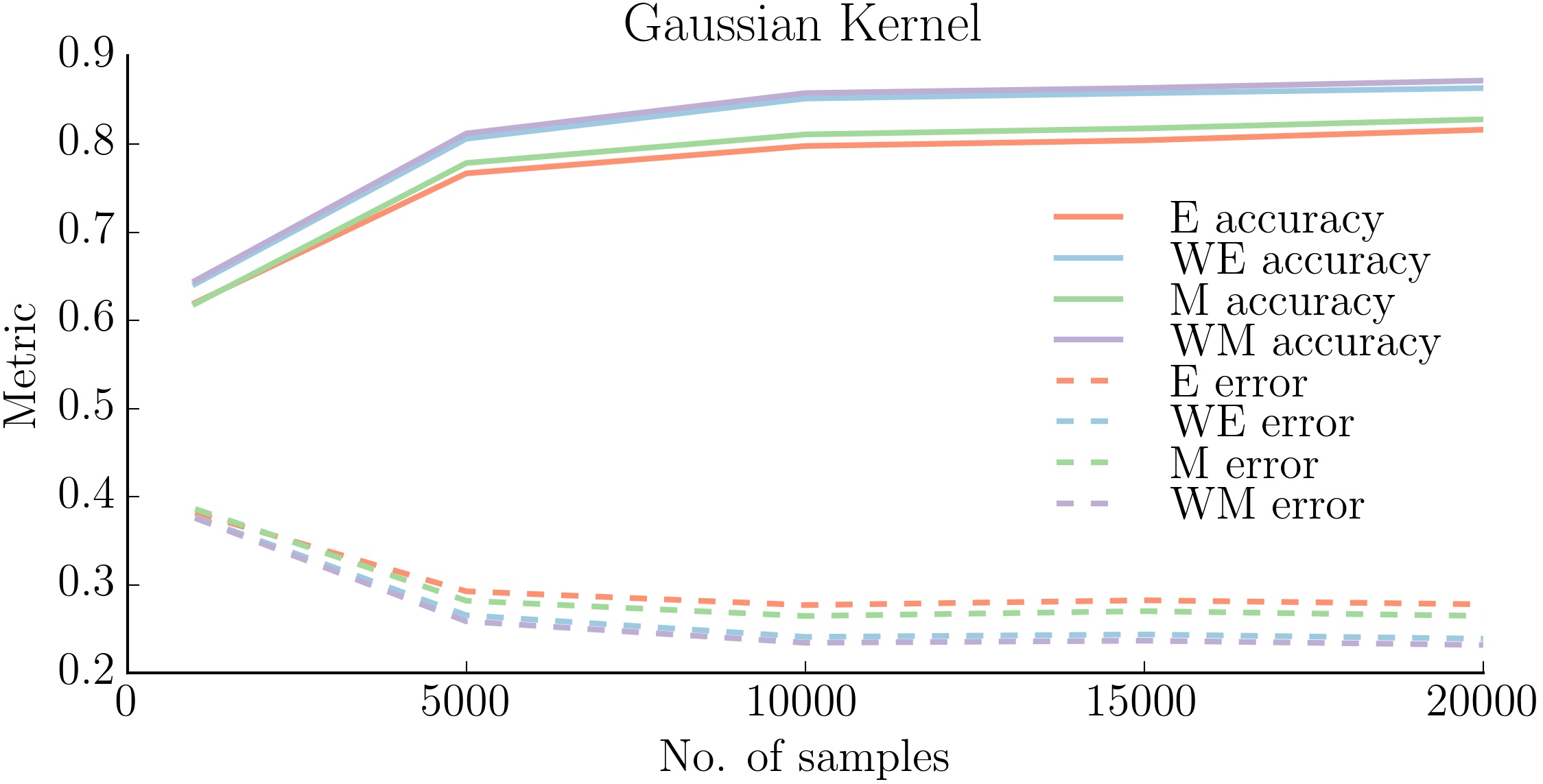}
            \subcaption{Gaussian Kernel}    
            \label{fig:gl_15}
            
        \end{minipage}
        \caption{The four figures show the results of the performance of a model using different methods (mentioned in the respective title) with different distance metric measures. (a) and (b) are plotted for $k=10$ while (c) and (d) for $r=1.5$.}
        \label{fig:res}
    \end{figure}
    
\section{Result}
Apart from the training size $N$, each of the four methods - Nearest Neighbours, Approximate Nearest Neighbours, Epanechnikov Kernel and Gaussian Kernel - has a parameter which influences the performance of the model. Number of neighbors $k$ is that parameter in the first two methods while the rest have the radius of sphere $r$. Out of all the set of parameters tried, the optimum performance is observed at $k = 10$ for the first two methods and $r = 1.5$ for the last two.

We analyzed the performance of the proposed weighted euclidean measure of computing distance metrics. The accuracy and average error of the model using the weighted euclidean measure were compared with that of the model invoking the typical euclidean measure. The results show that the weighted Euclidean measure performs better than the non-weighted one that too by a considerable margin for all the methods over the entire range of training set size. Our experiments demonstrated that it achieves higher accuracy than the latter one. In addition, the average error encountered in estimating C-space belief using this measure is less as compared to the Euclidean measure. 

Moreover, we analyzed the results of using scale invariant mahalanobis distance measure over euclidean measure in computing distance metrics. The model implementing the former measure results in higher accuracy and lower average error compared to the one which invokes euclidean measure. Figure \ref{fig:res} shows the results of using four different measures for evaluating distance metrics in different methods. 
 
\begin{figure}[t]
    \begin{minipage}{.5\textwidth}
        \includegraphics[width=\textwidth]{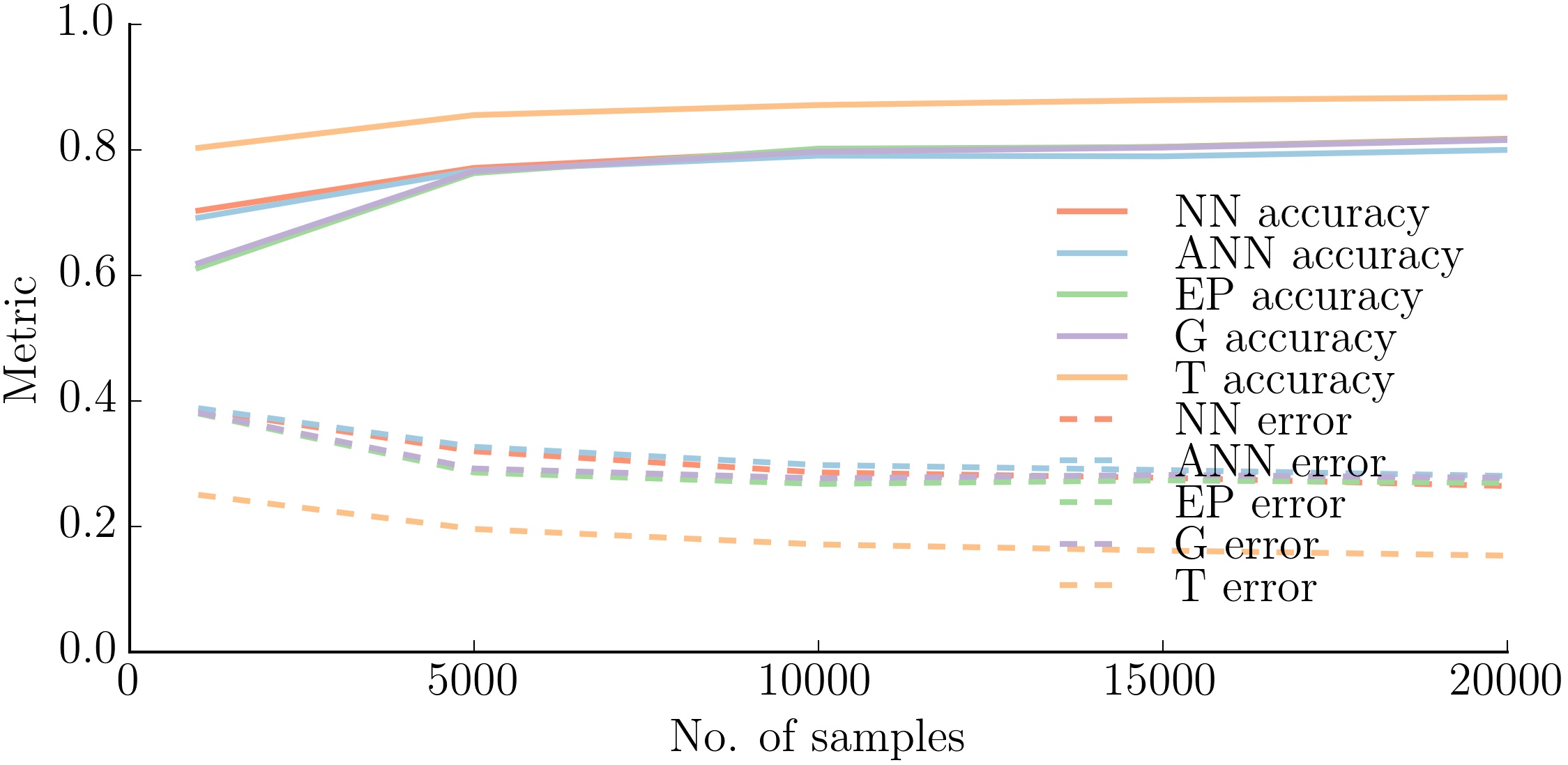}
        \subcaption{}
        \label{fig:all}
    \end{minipage}
    \begin{minipage}{.5\textwidth}
        \includegraphics[width=\textwidth]{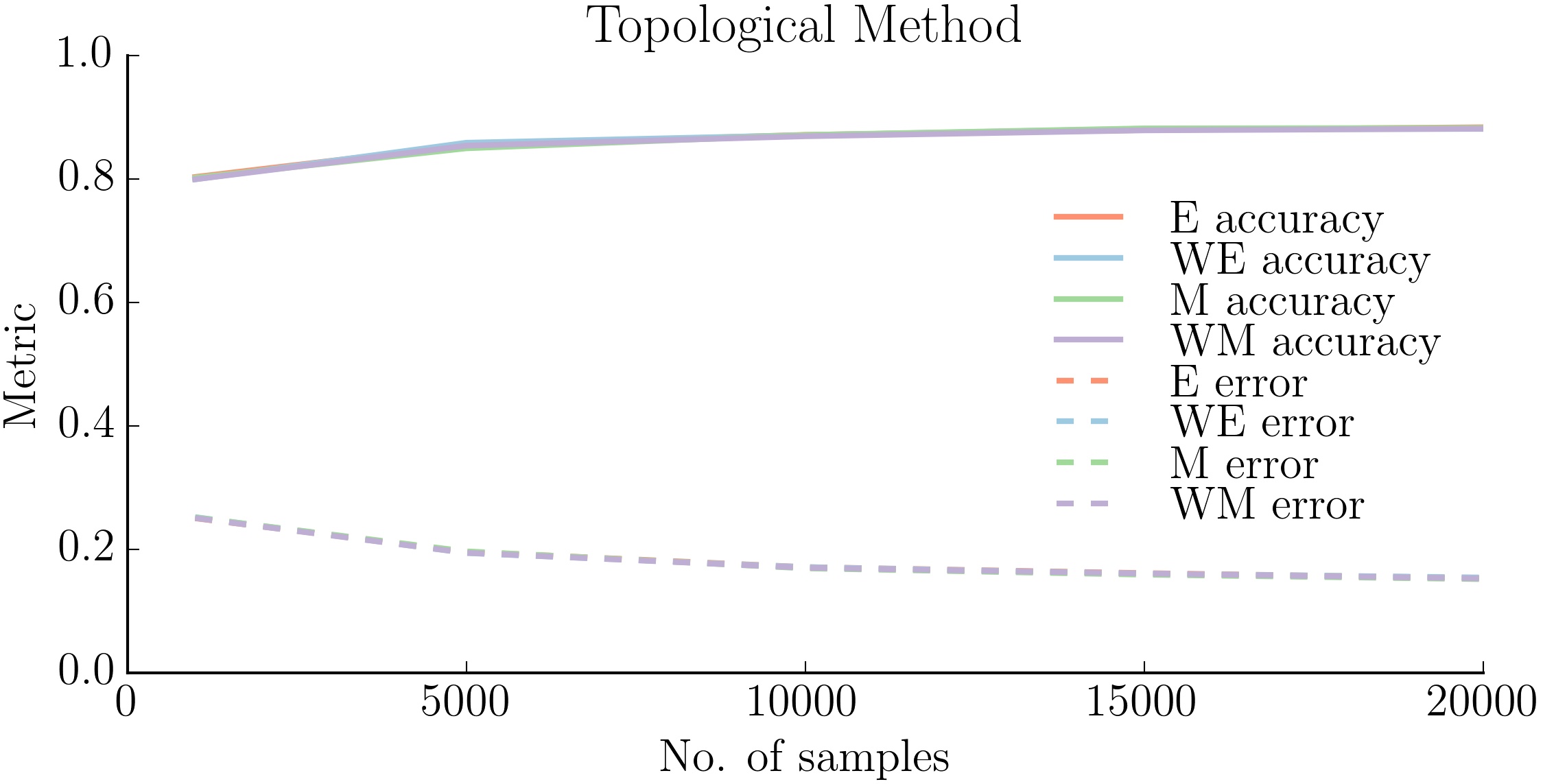}
        \subcaption{}
        \label{fig:dt}
    \end{minipage}
    \caption{(a) The accuracy and error of all the 5 methods are compared as a function of $N$. NN and ANN methods have $k=10$ neighbors while the kernel methods have radius $r=1.5$. In (b), the results of implementing various distance metric measures in the topological method are shown.}
\end{figure}

We compared the performance of our proposed topological method with the four other methods discussed above. The results are plotted in Figure \ref{fig:all}. As evident from the figure, the accuracy of our method is significantly higher than that of all the other baselines. In addition, the average error encountered in using topological methods is considerably lower than that in others. Figure \ref{fig:dt} presents the comparative analysis of four measures used to evaluate the distance metrics. 

\section{Discussion and Future Work}
In this work, we have proposed a diagonal matrix $\mathbf{W}$ to capture the unequal significance of arm joints. Weighted L2 norm with $\mathbf{W}$ as the weighting matrix is a more appropriate measure of the weight of a sample as compared to L2 norm. Currently, we have used the ability of a joint to move a reference point (end-effector in $C_{free}$ samples and collision center of mass in $C_{obs}$ samples) as a measure of the importance weight of the sample. However, the displacement of a single point in a 7 DOF arm does not capture the manipulability of the arm in the particular configuration. We believe that a better estimate of a joint weight can be the total volume swept by the arm due to small perturbation the joint.   

$\mathbf{W}$ is a diagonal matrix because of our method of estimating joint weights. Diagonal elements are the weight of the corresponding joint in the configuration. By the same analogy, non-diagonal terms represent the coupled weight of two joints. In other words, it is like correlation between two joints. It may be an interesting idea to reason about  the non-diagonal terms of the matrix and whether they can be anything non-zero. 

We have used Delaunay Triangulation as the topological structure of the projected space. The neighbours of a query are taken as the vertices of the circumscribing simplex. The results of this method are better than all the other ones that are used in the C-space itself. However, when the sampling density is high, the higher order structure is lost because of the presence of large number of simplices.  
To overcome these shortcomings, we plan to use neighbourhood graphs like Relative Neighbourhood Graph~\cite{toussaint1980relative} (RNG) as the topological structure. The motivation behind doing so is the fact that RNG of a set of points closely matches human perception of the shape of the set. The neighbours of a query can be taken as the vertices of the polygon enclosing the query.

\bibliographystyle{unsrt}
\bibliography{ref}

\end{document}